\documentclass{article}
\usepackage{spconf,amsmath,epsfig}

\providecommand{\doi}[1]{doi: {\footnotesize \href{http://dx.doi.org/#1}{\path{#1}}}}

\usepackage[pdftex=true,breaklinks=true,hidelinks=true,colorlinks=true,citecolor=blue]{hyperref}

\usepackage{multirow}
\usepackage[square,numbers]{natbib}

\title{Deep-Learning-based Change Detection\\ 
with Spaceborne Hyperspectral PRISMA data}
\name{J.F. Amieva$^{1*}$, A. Austoni$^1$\sthanks{These authors contributed equally.}, M.A. Brovelli$^1$, L. Ansalone$^2$, P. Naylor$^3$, 
F. Serva$^{2,3}$\sthanks{Now at the National Research Council, Rome, Italy}, B. Le Saux$^3$}
\address{\textsuperscript{1}Dipartimento di Ingegneria Civile e Ambientale,\\
         Politecnico di Milano, Milano I-20133, Italy\\
         \textsuperscript{2}Agenzia Spaziale Italiana,
         Via del Politecnico snc, Roma I-00133, Italy \\
         \textsuperscript{3}$\Phi$-lab, ESRIN, European Space Agency,
         Frascati I-00044, Italy\\
         }

\begin{document}
%
\maketitle
\begin{abstract}
Change detection (CD) methods have been applied to 
optical data for decades, 
while the use of hyperspectral data with a fine spectral resolution 
has been rarely explored.
CD is applied in several sectors, such as environmental 
monitoring and disaster management.
Thanks to the PRecursore IperSpettrale della Missione operativA (PRISMA), hyperspectral-from-space CD is now possible.
In this work, we apply standard and deep-learning (DL) CD methods to different targets, from natural to urban areas. 
We propose a pipeline starting from coregistration, 
followed by CD with a full-spectrum algorithm and by a DL
network developed for optical data.
We find that changes in vegetation and built environments
are well captured.
The spectral information is valuable to identify subtle changes and 
the DL methods are less affected by noise compared to the
statistical method, but atmospheric effects and the lack of reliable ground truth 
represent a major challenge to hyperspectral CD.

\end{abstract}

\section{Introduction}
\label{sec:intro} 

Change detection (CD) is the set of procedures used to identify changes between 
multiple images, generally acquired at different times, which has been applied 
with success in remote sensing data for several decades \cite{malila_change_1980}.
It is also known that different CD methods can produce different change maps
\cite{singh_digital_1989}, and that often expert assessment is required to
interpret and post-process results in a supervised fashion.

Deep learning (DL) methods have been gaining much attention as a tool for 
automatizing time-consuming CD tasks \cite{khelifi_deep_2020}.
Since CD involves identifying spatial features and their changes between 
two different dates, convolutional neural networks (NNs) have proven 
to be highly successful for CD in optical and radar data 
\cite{daudt_urban_2018, li_deep_2019}.

Unlike multispectral imagery, hyperspectral data provides very detailed 
information on the spectral characteristics of the sensed objects, allowing
for example to discriminate between different materials or retrieve biogeophysical 
parameters accurately \cite{qian_hyperspectral_21}. 
Historically, this kind of data has been collected from airborne platforms, 
while now multiple hyperspectral satellite missions are ongoing or planned,
opening a new era for their applications.

PRISMA (PRecursore IperSpettrale della Missione operativA) is a mission of the 
Italian Space Agency (ASI) acquiring hyperspectral data globally since 2019.
Further details on the mission are provided in Sec. \ref{sec:methods}.
To our knowledge, CD with PRISMA have been limited so far,
as recent works \cite{arjasakusuma_change_2022, shafique_ssvit_2023} considered 
only individual pairs of images.
In general, hyperspectral CD studies are challenged by the lack of suitable
data, often restricted to a few regions of the world  
\cite{fandino_stacked_2018, huang_hyperspectral_2019, wang_sub_2022} or without temporality \cite{fuchs_hyspecnet_2023}, with ground 
truth often unavailable for validation.
Contributions of this work comprise 1) information on a list of PRISMA image pairs enabling change detection; 2) an assessment of unsupervised statistical and DL methods for CD 
with PRISMA data to illustrate their limits and potentialities. 

\section{Data and Methods}
\label{sec:methods}
\subsection{PRISMA satellite data}
The PRISMA mission was launched in June 2019 and is one of the 
latest imaging spectroscopy missions for Earth observation 
\cite{lopinto_current_2021}.
The PRISMA satellite has a hyperspectral imaging spectrometer 
(30 m GSD) and a panchromatic camera (5 m GSD). 
The hyperspectral sensor covers the visible and near-infrared (VNIR: 400 – 1010 nm, 
66 bands) and the short-wave infrared (SWIR: 920 – 2505 nm, 174 bands) with a high 
spectral resolution, having a total of 240 bands.
The acquisitions have a swath of 30 km with a revisit time below 29 days, and since they are primarily on-demand, based on requests by 
registered users, multitemporal and cloud-free images are not 
always available. However, the satellite has near-global acquisition capabilities,
which provides potential coverage even in remote areas.
\begin{table}
\small
\caption{Details on the PRISMA acquisitions used.}
\label{table:acquisitions}
\centering
\begin{tabular}{|c|c|c|c|}
\hline \hline
\textbf{Location} & \textbf{Dates} & \textbf{Lat/Lon (deg)} \\
\hline \hline
Athens & 2020-07-22; 2022-07-05  & 37.94 / 23.95 \\
Beirut & 2020-08-23; 2022-06-26  & 33.86 / 35.55 \\
Hanging Rock & 2019-12-27; 2021-02-11 & -31.49 / 151.29\\
Java & 2021-04-21; 2021-07-17 &  -7.54 / 110.44\\
Lagos & 2020-11-13; 2022-01-22 &  6.44 / 3.39 \\
London & 2020-06-24; 2022-07-18 & 51.48 / -0.46 \\
Los Angeles & 2020-07-21; 2022-07-16 & 34.01 / -118.22 \\
Nalasopara & 2019-12-31; 2022-02-21 & 19.47 / 72.84 \\
Newark & 2020-04-15; 2022-04-22 &  40.72 / -74.2\\
Rome & 2020-08-06; 2022-06-15 & 41.86 / 12.26 \\
Shanghai & 2021-04-09; 2022-04-09  & 31.35 / 121.6 \\
\hline \hline
\end{tabular}
\end{table}
For this study, we use atmospheric-corrected and geocoded surface reflectance 
(L2D) data provided by ASI.
An additional coregistration step is used to reduce shifts between 
images down to the pixel level. 

Eleven pairs of PRISMA acquisitions with low cloud coverage are selected for our analysis.
They are chosen to ensure global representativeness, sampling of different land cover 
states (e.g., rural, urban, or mixed), and consistent timing
to reduce seasonal variations. Further information as acquisition
times and coordinates is provided in Table \ref{table:acquisitions}.

\subsection{Preprocessing and CD methods}
\begin{figure}
  \includegraphics[width=.87\columnwidth]{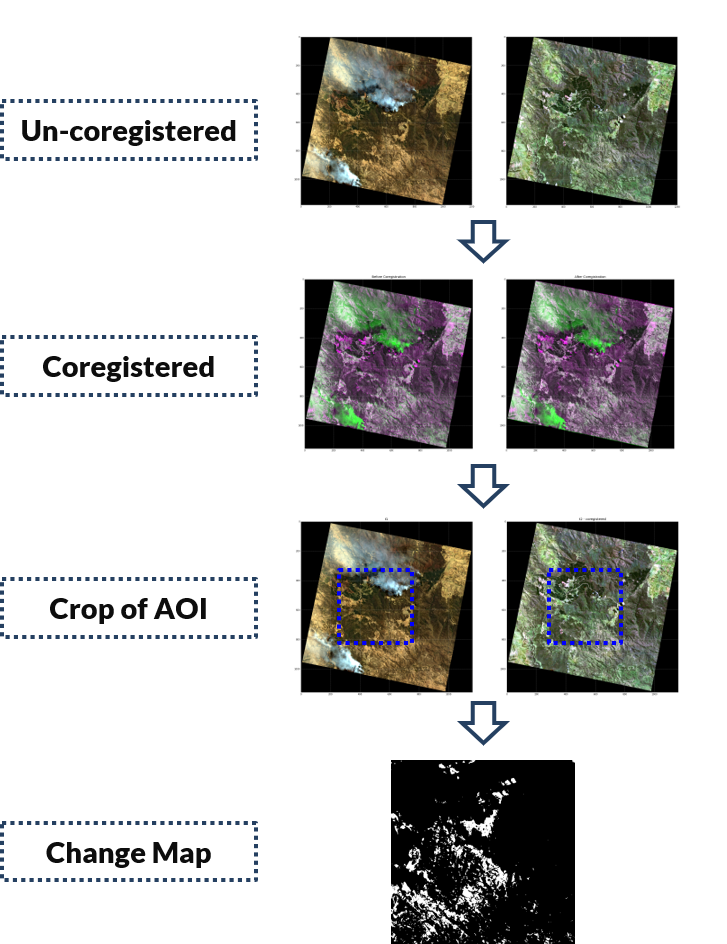}
  \caption{\label{fig:wflow} Processing workflow example:
  RGB images and AOI (blue) (rows 1-3), 
  CD map (row 4 - white: change, black: no change).
  Data processed under license (\copyright ASI).}\vspace{-0.5cm}
\end{figure}

PRISMA image pairs are \textbf{co-registered} with the \emph{GeFolki} software
\cite{brigot_adaptation_2016}, using the red band of PRISMA images, with similar results confirmed for other band selections.
The `before' ($b$) image is used as the target, and the `after' 
($a$) is the moving image.
The optical flow derived with this method for the overlapping area 
is then used to correct all the bands of the moving image to better 
match the target, generally resulting in shifts below 5 pixels 
for complex terrains.
Finally, square patches with size $512\times512$ for the area of 
interest (AOI) are extracted and used to derive binary CD maps 
(Fig. \ref{fig:wflow}).

\begin{figure}
  \includegraphics[width=\columnwidth]{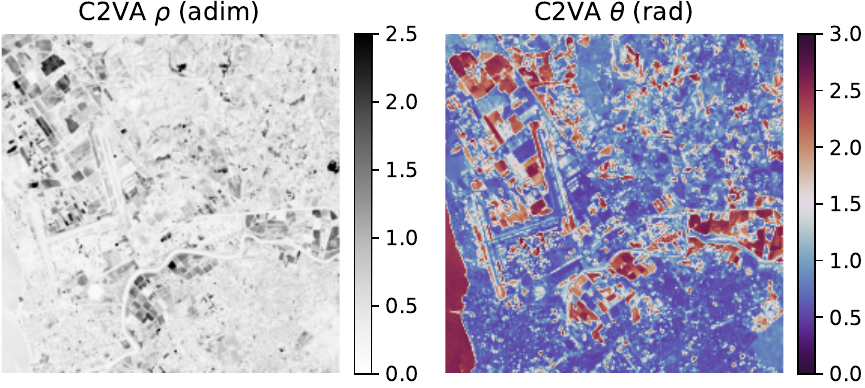}
  \caption{Illustration of C2VA magnitude and angle 
  obtained for the Rome image pair.
  Data processed under license (\copyright ASI).}
\label{fig:c2va_ex}
\end{figure}

Compressed \textbf{change vector analysis} (C2VA) is an unsupervised method
\cite{bovolo_framework_2012} that fully exploits 
multispectral image information.
Similar to the standard CVA \cite{malila_change_1980}, for each pixel a change magnitude ($\rho$) 
and phase angle ($\theta$) are calculated for the before and after
images for $B$ spectral bands, as
$\rho=\sqrt{\sum_{k=1}^B (X_{k,a}-X_{k,b})^2}$. 
The phase angle is estimated from an arbitrary reference vector, e.g.
$X_{ref}=({\sqrt{B}}/{B} , ... , {\sqrt{B}}/{B})$, and can be 
used to identify coherent changes.
Examples of magnitude and phase angle from C2VA for one scene are 
reported in Fig. \ref{fig:c2va_ex}.
Notably changes in agricultural fields (upper left areas) or the mining area
on the right have a similar phase, suggesting their common nature. 
For this work we focus on the magnitude information, making binary change 
maps based on a 90$^{th}$ percentile threshold calculated for each pair.

\textbf{DeepCVA} (DCVA)~\cite{saha2019unsupervised} is a method for generating a pixel wise hyper vector $G$ from the difference of a chosen subset of layers $L$ from a pre-trained deep NN and performing unsupervised CD. 
The chosen subset is given apriori and is denoted by $L \subset [1:N]$ where $N$ is the number of layers. 
We denote by $F$ the NN composed of layers $F^l$ with $l\in [1:N]$.
To simplify the notation, we denote by $F_i^l$ the feature layer of $X_i$ at layer $F^l$ with $i \in \{a,b\}$. 
$G$ is generated from the difference of extracted deep representations. 
In particular, we compute the difference for each given feature-layer $l\in L$: $\delta_l = F^l_b - F^l_a$ and concatenate each $\delta_l$ to obtain $G$.
To limit the size of $G$, we refine the number of selected features at each layer of $L$ by only retaining a certain percentile of features from $\delta_l$.
For each layer $l$, we perform a pixel-wise spatial clustering and keep for each cluster the features with the highest variance within the top percentile.
CD is then computed at a pixel-wise level, where a change is reported if $||G|| > \mathcal{T}$ where $\mathcal{T}$ is a given threshold.
For the computation of $\mathcal{T}$, we use a local adaptive thresholding method (DCVA Ada) or Otsu's method (DCVA Otsu).
We use a network pre-trained \cite{saha2019unsupervised} on visible/infrared bands (RGBIR).
\section{Results and discussion}
\label{sec:pagestyle}
\begin{figure*}
\footnotesize
\label{fig:res_shanghai}
\begin{minipage}[b]{.16\linewidth}
  \includegraphics[width=\columnwidth]{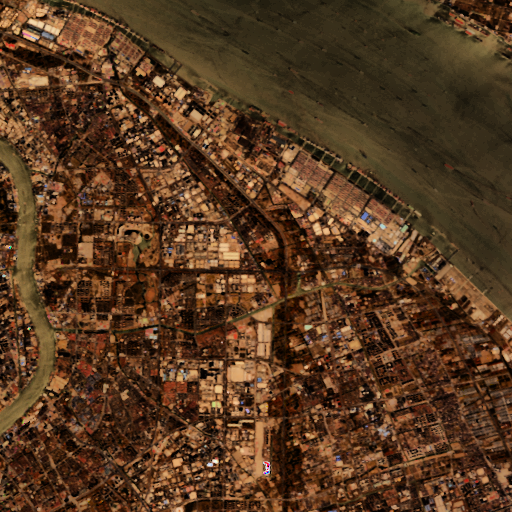}
  \centerline{RGB before}
\end{minipage}
\hfill
\begin{minipage}[b]{0.16\linewidth}
  \includegraphics[width=\columnwidth]{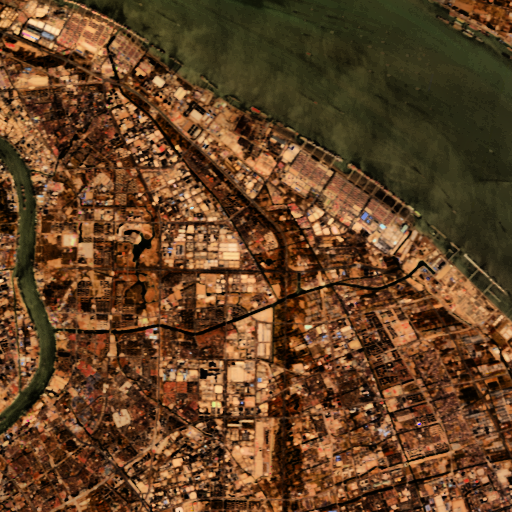}
  \centerline{RGB after}
\end{minipage}
\hfill
\begin{minipage}[b]{0.16\linewidth}
  \includegraphics[width=\columnwidth]{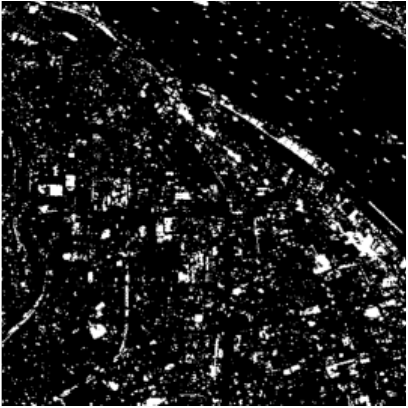}
  \centerline{C2VA}
\end{minipage}
\hfill
\begin{minipage}[b]{.16\linewidth}
  \includegraphics[width=\columnwidth]{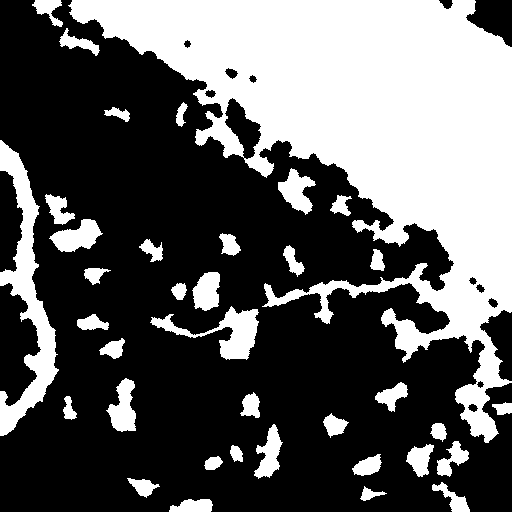}
 \centerline{DCVA Otsu}
\end{minipage}
\hfill
\begin{minipage}[b]{.16\linewidth}
  \includegraphics[width=\columnwidth]{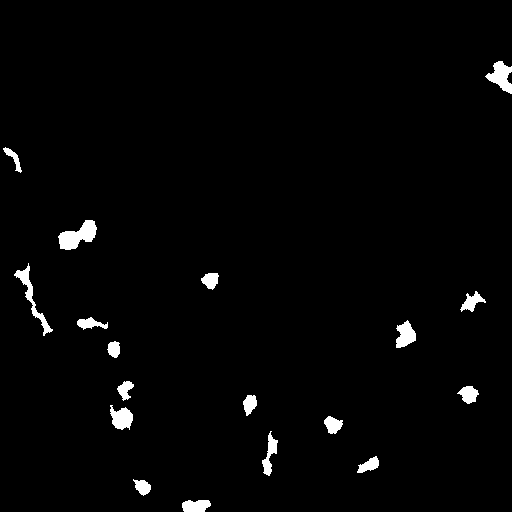}
 \centerline{DCVA Ada L2,5}
\end{minipage}
\hfill
\begin{minipage}[b]{.16\linewidth}
  \includegraphics[width=\columnwidth]{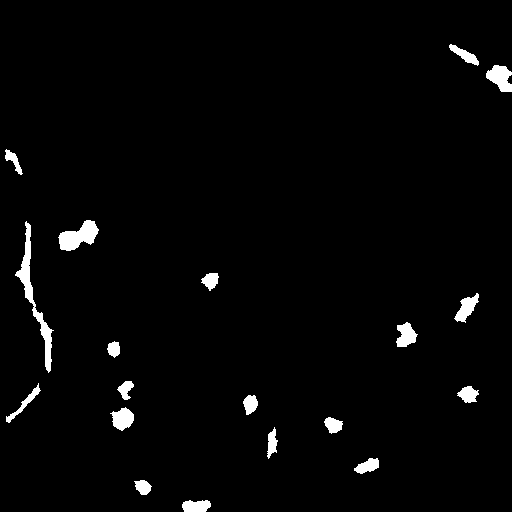}
  \centerline{DCVA Ada L2,5,8,10,11,23}
\end{minipage}
\caption{Image pair for the Shanghai scene and change maps
by different methods.
Data processed under license (\copyright ASI).}
\label{fig:res}
\end{figure*}

\begin{figure*}
\footnotesize
\label{fig:rome_panel}
\begin{minipage}[b]{.16\linewidth}
  \includegraphics[width=\columnwidth]{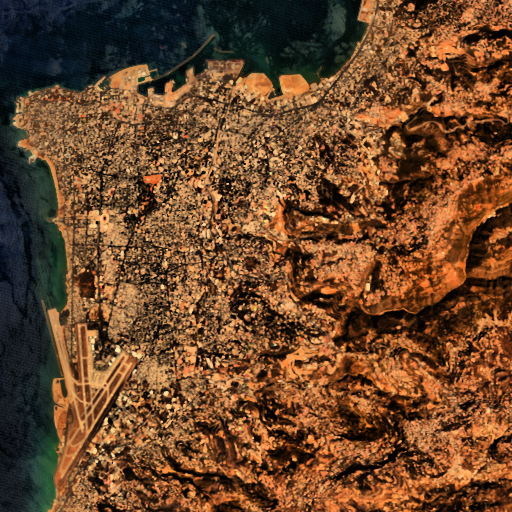}
  \centerline{RGB before}
\end{minipage}
\hfill
\begin{minipage}[b]{0.16\linewidth}
  \includegraphics[width=\columnwidth]{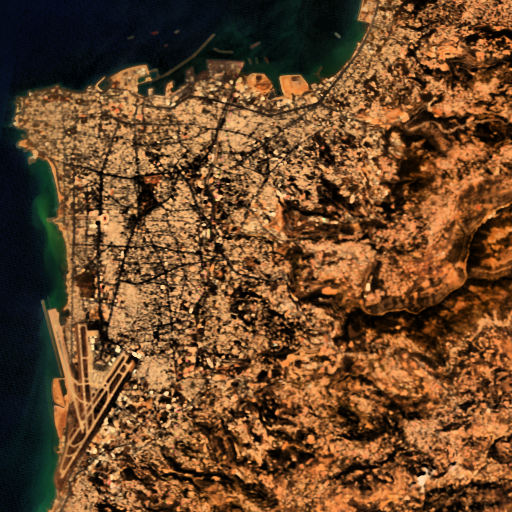}
  \centerline{RGB after}
\end{minipage}
\hfill
\begin{minipage}[b]{0.16\linewidth}
  \includegraphics[width=\columnwidth]{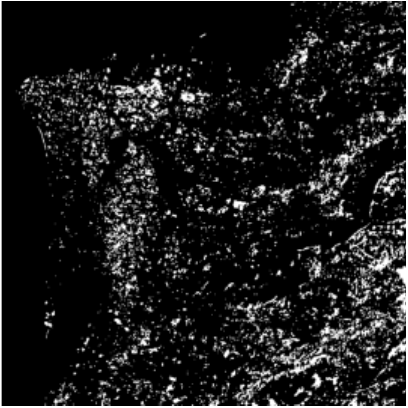}
  \centerline{C2VA}
\end{minipage}
\hfill
\begin{minipage}[b]{.16\linewidth}
  \includegraphics[width=\columnwidth]{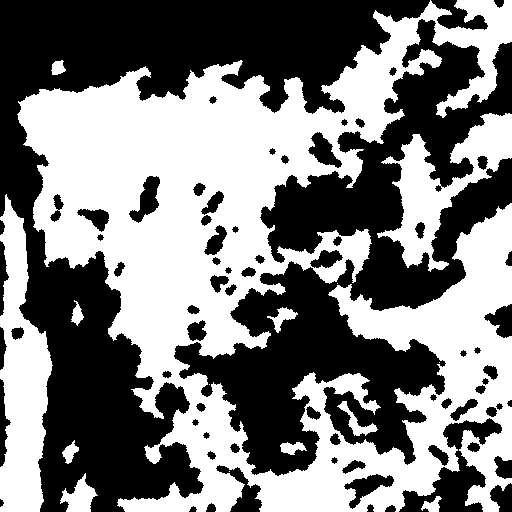}
 \centerline{DCVA Otsu}
\end{minipage}
\hfill
\begin{minipage}[b]{.16\linewidth}
  \includegraphics[width=\columnwidth]{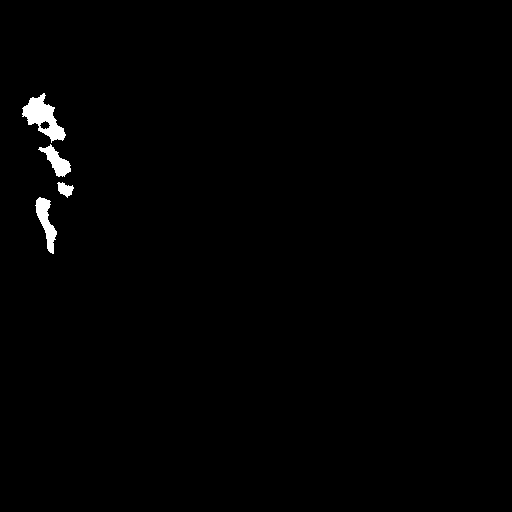}
 \centerline{DCVA Ada L2,5}
\end{minipage}
\hfill
\begin{minipage}[b]{.16\linewidth}
  \includegraphics[width=\columnwidth]{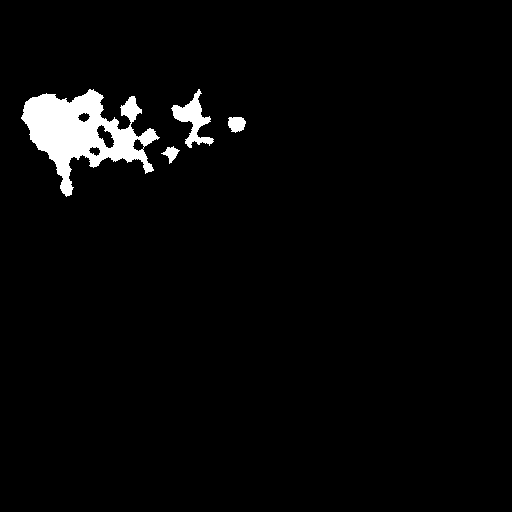}
  \centerline{DCVA Ada L2,5,8,10,11,23}
\end{minipage}
\caption{As in Fig. \ref{fig:res_shanghai}, but for Beirut.}
\label{fig:res_beirut}
\end{figure*}

An overview of CD results for all the pairs of PRISMA acquisitions is 
provided in Table \ref{tab:cd_perc}, comparing
different scenes and methods.
By construction, 10\% of 
the image is marked as change for C2VA.
DCVA with Otsu thresholding method is prone to overestimating the amount of changes. 
This is likely due to the fact that 
the difference histograms produced are not bimodal and therefore
binarization is affected by noise or equalization issues 
(e.g., due to clouds). 
The estimates of DCVA Ada methods are much more conservative,
with little or no change detected for urban scenes, 
such as Athens or Los Angeles.
It should be noted that the network was pre-trained on 
high-resolution optical imagery, and the coarse spatial resolution
of PRISMA may not be fully suitable for the task.
The amount of changes identified by the network is not 
trivially related to the number of layers selected.
For example in the Beirut scene, increasing the number of layers induces more detected changes,
but fewer changes for the Shanghai pair.
Moreover, in the case of Lagos, the presence of haze or 
smog in one image (not shown) leads to the smallest amount of changes
detected overall.
\begin{table}
\center
\small
\caption{Number of changed pixels (in \%). 
($i$) indicates the number of layers. (1) is layers $[2,5]$, (2) is $[2,5,8,10]$ and (3) is $[2,5,8,10,11,23]$. 
DCVA Otsu is with (2).}
\label{tab:cd_perc}
\begin{tabular}{|c|c|c|c|c|c|c|}
\hline \hline
\multirow{2}{*}{\textbf{Location}} & \multirow{2}{*}{\textbf{Type}} & \multirow{2}{*}{\textbf{C2VA}} & \textbf{DCVA} & \multicolumn{3}{c|}{\textbf{DCVA Ada}} \\ \cline{5-7}
& &  & \textbf{Otsu} & (1) & (2) & (3) \\ \hline \hline
Athens & urban & 10 & 37.4 & 0.0 & 0.1 & 0.3 \\
Beirut & mix & 10 & 49.4 & 0.8 & 2.8 & 3.0 \\
Hanging Rock & rural & 10 & 39.4 & 0.6 & 1.5 & 1.1 \\
Java & rural & 10 & 22.0 & 0.7 & 2.7 & 2.7 \\
Lagos & urban &  10 & 4.2 & 2.2 & 2.6 & 2.3 \\
London & urban & 10 & 23.4 & 3.2 & 4.7 & 4.9 \\
Los Angeles & urban & 10 & 34.3 & 0.1 & 0.1 & 0.0 \\
Nalasopara & rural & 10 & 29.4 & 2.7 & 4.8 & 4.5 \\
Newark & urban & 10 & 16.6 & 2.1 & 2.7 & 3.4 \\
Rome & mix & 10 & 26.4 & 5.0 & 3.9 & 3.7 \\
Shanghai & urban & 10 & 37.0 & 2.0 & 2.3 & 1.6 \\
\hline \hline
\end{tabular}
\end{table}
To visualize the contrast between the different 
methods and sensitivity to hyperparameters, we
report a comparison of CD masks and corresponding
RGB images in Fig.~\ref{fig:res_shanghai} and \ref{fig:res_beirut}.
These images illustrate how DCVA Otsu tends to mark 
water surfaces as changed, likely due to their different optical 
characteristics.
Detected changes by the adaptive methods are a subset of Otsu's
method, and more changes are picked up when more layers are
considered.
This may be explained by noting that a small number of layers is 
more suitable for homogeneous scenes. 
For the Beirut scene, in the most conservative setting (layer 2,5), changes are detected in the city centre, near the August 2020 explosion site.
An interesting feature in Fig.~\ref{fig:res_shanghai}
is that C2VA can identify ships in the upper right
portion of the image, while they are missing when the change is 
identified in the latent space.
For specific applications, such as ship detection, 
it would be important to tune the algorithms with real data as fine-grained details can be missing in the latent space.
Large-scale land-use changes are however identified by both methods.
In general, changes identified by C2VA appear noisy
in urban scenes but are fairly consistent when their extent is larger,
without the sensitivity of the DCVA Otsu for water bodies.
In the absence of reliable ground-truth data, it is however
hard to judge which method is performing better.


\section{Conclusions and perspectives}
In this work, we present a comparison of statistical and 
DL methods for performing change detection on novel hyperspectral 
satellite data (from PRISMA).
This is the first study to address hyperspectral-from-space with deep learning on a varied portfolio of changes, as operational hyperspectral satellite missions have only recently been launched.
For this task, we selected eleven pairs of acquisitions.
The areas were selected to represent the heterogeneity commonly found in EO data, including areas with clouds or ship traffic.

We find that larger scale changes in natural and urban scenes are
successfully identified by the proposed methods.
Sensitivity to terrain conditions and atmospheric effects is also noted.
The moderate spatial resolution of PRISMA complicates the application of a 
pre-trained network using only four bands. In this case, results are 
very sensitive to the number of layers adopted.
Training from scratch with a larger multispectral dataset would likely improve results.

Our work shows the potential of hyperspectral data for CD tasks. 
The lack of reliable ground truth data complicates the assessment of the 
different methods, but subjective evaluation indicates that the use of 
threshold-based methods is not always successful.
The availability of datasets with high-quality ground truth labels 
would be useful for many applications, including the development of DL models and for semantic CD.
We aim to release a public dataset for this purpose, as this is now possible thanks 
to current and new missions such as CHIME \cite{nieke_status_2019}.


\small
\bibliographystyle{plainnat}
\bibliography{biblio.bib}

\begin{thebibliography}{17}
\providecommand{\natexlab}[1]{#1}
\providecommand{\url}[1]{\texttt{#1}}
\expandafter\ifx\csname urlstyle\endcsname\relax
  \providecommand{\doi}[1]{doi: #1}\else
  \providecommand{\doi}{doi: \begingroup \urlstyle{rm}\Url}\fi

\bibitem[Arjasakusuma et~al.(2022)]{arjasakusuma_change_2022}
S.~Arjasakusuma et~al.
\newblock Change {Detection} {Analysis} using {Bitemporal} {PRISMA}
  {Hyperspectral} {Data}: {Case} {Study} of {Magelang} and {Boyolali}
  {Districts}, {Central} {Java} {Province}, {Indonesia}.
\newblock \emph{J. Ind. Soc. Rem. Sens.}, 2022.
\newblock \doi{10.1007/s12524-022-01566-z}.

\bibitem[Bovolo et~al.(2012)Bovolo, Marchesi, and
  Bruzzone]{bovolo_framework_2012}
F.~Bovolo, S.~Marchesi, and L.~Bruzzone.
\newblock A {Framework} for {Automatic} and {Unsupervised} {Detection} of
  {Multiple} {Changes} in {Multitemporal} {Images}.
\newblock \emph{IEEE TGRS}, 50\penalty0 (6):\penalty0 2196--2212, 2012.
\newblock \doi{10.1109/TGRS.2011.2171493}.

\bibitem[Brigot et~al.(2016)]{brigot_adaptation_2016}
G.~Brigot et~al.
\newblock Adaptation and {Evaluation} of an {Optical} {Flow} {Method} {Applied}
  to {Coregistration} of {Forest} {Remote} {Sensing} {Images}.
\newblock \emph{IEEE JSTARS}, 9\penalty0 (7):\penalty0 2923--2939, 2016.
\newblock \doi{10.1109/JSTARS.2016.2578362}.

\bibitem[{Caye Daudt} et~al.(2018)]{daudt_urban_2018}
R.~{Caye Daudt} et~al.
\newblock Urban change detection for multispectral earth observation using
  convolutional neural networks.
\newblock In \emph{IEEE IGARSS}, July 2018.

\bibitem[Fuchs and Demir(2023)]{fuchs_hyspecnet_2023}
M.~H.~P. Fuchs and B.~Demir.
\newblock {HySpecNet-11k}: A large-scale hyperspectral dataset for benchmarking
  learning-based hyperspectral image compression methods.
\newblock \emph{arXiv:2306.00385 [cs.CV]}, 2023.

\bibitem[Huang et~al.(2019)Huang, Yu, and Feng]{huang_hyperspectral_2019}
F.~Huang, Y.~Yu, and T.~Feng.
\newblock Hyperspectral remote sensing image change detection based on tensor
  and deep learning.
\newblock \emph{J. of Vis. Comm. Image Repr.}, 58:\penalty0 233--244, 2019.
\newblock \doi{10.1016/j.jvcir.2018.11.004}.

\bibitem[Khelifi and Mignotte(2020)]{khelifi_deep_2020}
L.~Khelifi and M.~Mignotte.
\newblock Deep {Learning} for {Change} {Detection} in {Remote} {Sensing}
  {Images}: {Comprehensive} {Review} and {Meta}-{Analysis}.
\newblock \emph{IEEE Access}, 8:\penalty0 126385--126400, 2020.
\newblock \doi{10.1109/ACCESS.2020.3008036}.

\bibitem[Li et~al.(2019)]{li_deep_2019}
Y.~Li et~al.
\newblock A deep learning method for change detection in synthetic aperture
  radar images.
\newblock \emph{IEEE TGRS}, 57\penalty0 (8):\penalty0 5751--5763, 2019.
\newblock \doi{10.1109/TGRS.2019.2901945}.

\bibitem[Lopinto et~al.(2021)]{lopinto_current_2021}
E.~Lopinto et~al.
\newblock Current {Status} and {Future} {Perspectives} of the {PRISMA}
  {Mission} at the {Turn} of {One} {Year} in {Operational} {Usage}.
\newblock In \emph{2021 {IGARSS}}, pages 1380--1383, 2021.
\newblock \doi{10.1109/IGARSS47720.2021.9553301}.

\bibitem[López-Fandiño et~al.(2018)]{fandino_stacked_2018}
J.~López-Fandiño et~al.
\newblock Stacked autoencoders for multiclass change detection in hyperspectral
  images.
\newblock In \emph{IGARSS 2018}, pages 1906--1909, 2018.
\newblock \doi{10.1109/IGARSS.2018.8518338}.

\bibitem[Malila(1980)]{malila_change_1980}
M.~W. Malila.
\newblock Change vector analysis: an approach for detecting forest changes with
  landsat.
\newblock \emph{Proc. 6th Ann. Sympos. Mach. Proc. Rem. Sens. Data, Purdue
  Univ.}, pages 326--335, 1980.

\bibitem[Nieke and Rast(2019)]{nieke_status_2019}
J.~Nieke and M.~Rast.
\newblock Status: {Copernicus} {Hyperspectral} {Imaging} {Mission} {For} {The}
  {Environment} ({CHIME}).
\newblock In \emph{{IGARSS} 2019}, pages 4609--4611, 2019.
\newblock \doi{10.1109/IGARSS.2019.8899807}.

\bibitem[Qian(2021)]{qian_hyperspectral_21}
S.-E. Qian.
\newblock Hyperspectral {Satellites}, {Evolution}, and {Development} {History}.
\newblock \emph{IEEE JSTARS}, 14:\penalty0 7032--7056, 2021.
\newblock \doi{10.1109/JSTARS.2021.3090256}.

\bibitem[Saha et~al.(2019)Saha, Bovolo, and Bruzzone]{saha2019unsupervised}
S.~Saha, F.~Bovolo, and L.~Bruzzone.
\newblock Unsupervised deep change vector analysis for multiple-change
  detection in {VHR} images.
\newblock \emph{IEEE TGRS}, 57\penalty0 (6):\penalty0 3677--3693, 2019.

\bibitem[Shafique et~al.(2023)]{shafique_ssvit_2023}
A.~Shafique et~al.
\newblock {SSViT}-{HCD}: {A} {Spatial} {Spectral} {Convolutional} {Vision}
  {Transformer} for {Hyperspectral} {Change} {Detection}.
\newblock \emph{IEEE JSTARS}, pages 1--20, 2023.
\newblock \doi{10.1109/JSTARS.2023.3251646}.

\bibitem[Singh(1989)]{singh_digital_1989}
A.~Singh.
\newblock {Digital} change detection techniques using remotely-sensed data.
\newblock \emph{Int. J. Rem. Sens.}, 10\penalty0 (6):\penalty0 989--1003, 1989.
\newblock \doi{10.1080/01431168908903939}.

\bibitem[Wang et~al.(2022)Wang, Wan, and Bruzzone]{wang_sub_2022}
L.~Wang, L.~Wan, and L.~Bruzzone.
\newblock A {Sub}-{Pixel} {Convolution}-{Based} {Residual} {Network} for
  {Hyperspectral} {Image} {Change} {Detection}.
\newblock In \emph{{IGARSS} 2022}, pages 1059--1062, 2022.
\newblock \doi{10.1109/IGARSS46834.2022.9884805}.

\end{thebibliography}

\end{document}